\documentclass[runningheads]{llncs}
\usepackage{graphicx}
\usepackage{comment}
\usepackage{amsmath,amssymb} 
\usepackage{color}
\usepackage{subfigure}
\usepackage{caption}
\usepackage{xspace}
\usepackage{booktabs,tabularx}
\usepackage{multirow}

\begin{document}
\pagestyle{headings}
\mainmatter
\def\ECCVSubNumber{2044}  

\title{Defocus Blur Detection via Depth Distillation} 

\titlerunning{Defocus Blur Detection via Depth Distillation}
\author{Xiaodong Cun \and Chi-Man Pun }

\authorrunning{X. Cun and C.-M. Pun}

\institute{University of Macau, Macau, China \\
\email{\{yb87432,cmpun\}@umac.mo}}

\maketitle

\newcolumntype{R}{>{\raggedleft\arraybackslash}X}%

\begin{abstract}
Defocus Blur Detection~(DBD) aims to separate in-focus and out-of-focus regions from a single image pixel-wisely. This task has been paid much attention since bokeh effects are widely used in digital cameras and smartphone photography. However, identifying obscure homogeneous regions and borderline transitions in partially defocus images is still challenging. To solve these problems, we introduce depth information into DBD for the first time. When the camera parameters are fixed, we argue that the accuracy of DBD is highly related to scene depth. 
Hence, we consider the depth information as the approximate soft label of DBD and propose a joint learning framework inspired by knowledge distillation. In detail, we learn the defocus blur from ground truth and the depth distilled from a well-trained depth estimation network at the same time. Thus, the sharp region will provide a strong prior for depth estimation while the blur detection also gains benefits from the distilled depth.
Besides, we propose a novel decoder in the fully convolutional network~(FCN) as our network structure. In each level of the decoder, we design the Selective Reception Field Block~(SRFB) for merging multi-scale features efficiently and reuse the side outputs as Supervision-guided Attention Block~(SAB). Unlike previous methods, the proposed decoder builds reception field pyramids and emphasizes salient regions simply and efficiently. Experiments show that our approach outperforms 11 other state-of-the-art methods on two popular datasets. Our method also runs at over 30 fps on a single GPU, which is 2x faster than previous works. The code is available at: https://github.com/vinthony/depth-distillation

\keywords{Defocus Blur Detection, Attention Module, Knowledge Distillation}

\end{abstract}

\section{Introduction}
Defocus blur, which is also called the bokeh effect in photography, has been widely used in everyday photos. The focus region emphasizes the salient object while the out-of-focus blur can protect the privacy of people appearing in the photo. Moreover, detecting this kind of blur is also crucial since the detected defocus region could be potentially useful in performing tasks. Such tasks include auto-refocus\cite{bae2007defocus}, salient object detection~\cite{jiang2013salient} and image retargeting~\cite{karaali2016image}.

Since DBD has a long history in computer vision~\cite{Shi:2014vk,Park:2017uq,yi2016lbp,Shi:2014vk,Golestaneh:2017tv}, traditional methods focus on designing novel hand-crafted features such as the gradient~\cite{Golestaneh:2017tv,xu2017estimating} or the frequency domain features~\cite{Shi:2014vk,shi2015just,yi2016lbp}. However, these methods extract limited features and lack high-level semantic information. Thus, if the scene is complex, it is hard to discriminate the defocus region by particular features.

\begin{figure}[t]
\includegraphics[width=\columnwidth]{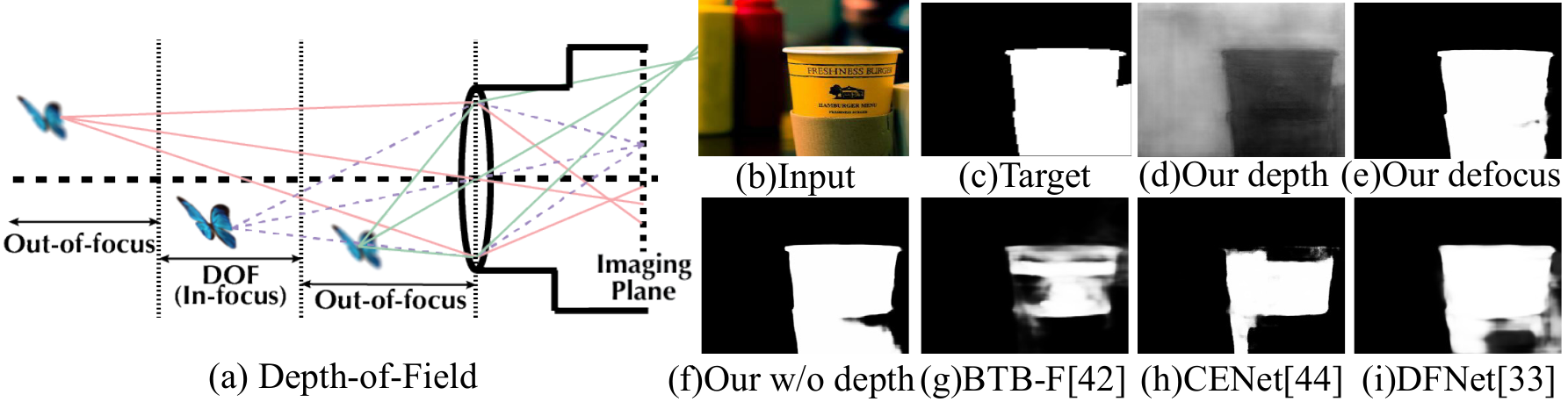}
\caption{We first leverage depth into DBD since predicting defocus blur is similar to estimating the \textit{Depth-of-Field}(DOF) from a partial defocus image as in (a). By involving depth in DBD network, our method assumes that the depths in DOF regions are similar(d) and that the region with more similar depths might be part of DOF/out-of-focus region as well(e). Thus, we obtain more accurate results than other DBD methods(f)-(i).}
\label{fig:dof}
\end{figure}

 Recently, deep learning-based methods have shown superior performance in various computer vision tasks as well as defocus blur detection. For example, Park~\textit{et al.}~\cite{Park:2017uq} train a CNN to classify the sharpness of each local patch in an image. Deeper fully convolutional methods~\cite{Zhao_2019_CVPR,zhao2018defocus,zhao2019defocus,tang2019defusionnet} have been proposed by regarding the DBD as scene segmentation. Although these methods emphasize the importance of image scales in DBD, they are still considering DBD from a 2D perspective and rely solely on the power of the datasets and neural network.
 
 In this paper, we start from the cause of defocus blur in the photography. As shown in Fig.~\ref{fig:dof}(a), the sharp focus region, also called the depth-of-field(DOF\cite{wiki:xxx}), is formed because the camera only images clear photo in a certain depth range\footnote{We simplify this model by ignoring the influence of camera parameters since we can only obtain a 2D RGB image in the dataset.}. When the light waves intersect behind or in front of the imaging plane~(red and green lines in Fig.~\ref{fig:dof}(a)), the area they originated from will be blurred in final image. Since the homogeneous region in DBD often includes multiple objects and since it is difficult to be detected by edges or semantic features, the distance between the camera and scene objects~(depth) provides a strong prior for classification. However, the unconstrained depth estimation is an ill-posed problem. To evaluate on currently available DBD datasets and provide fair comparison with previous methods, we propose depth distillation by using a pre-trained network~\cite{chen2016single} as regularization and learn the defocus map simultaneously.  
 In addition, we design a Supervision-guided Attention Block~(SAB) for re-weighting the learned features based on each level of side outputs.
 Finally, the blur confidence is relative, which means \textit{a sharp patch can be regarded as blurry when we enlarge it and vice versa}. Although previous methods~\cite{zhao2018defocus,zhao2019defocus,tang2019defusionnet} have discussed it by multi-stream or cross-layer fusion networks, we consider it in an efficient way by designing Selective Reception Field Block~(SRFB) in each decoder. Our block extracts larger reception fields to build richer feature pyramids and uses a global selective attention to weight the importance of useful features.  
By involving depth estimation into DBD and the proposed blocks, our network outperforms other methods on the defocus detection. As shown in Fig.~\ref{fig:dof} (b)-(i), previous methods for DBD are sensitive to color, while in our network, DBD and depth estimation tasks build on each other and predict the results perfectly. 

We summarize the contributions of this paper as follows:
 
\begin{itemize}
  \item To the best of our knowledge, this is the first attempt to introduce depth information in DBD and distill the knowledge of pre-trained depth model as regularization of DBD network.
  \item In each decoder of our framework, we design the Supervision-guided Attention Block~(SAB), which reuses the side depth and defocus map for spatial attention. Considering the sensitivity of scale, we also design the Selective Reception Fields Block~(SRFB) to extract multiple reception field features.
  \item We conduct the experiments on two popular DBD benchmarks with 11 state-of-the-art methods~(7 from DBD and 4 from related tasks). The results show that our proposed method can achieve much better results.
\end{itemize}

\section{Related Works}
\noindent\textbf{Traditional methods}
Out-of-focus and DOF regions have significant visible differences in sharpness. Thus, traditional DBD methods are designed based on identifiable hand-crafted features such as gradient or edge representation~\cite{karaali2017edge,Shi:2014vk}. For example, Yi~\textit{et al.}~\cite{yi2016lbp} use local binary patterns as focus sharpness metric. Shi~\textit{et al.}~\cite{shi2015just} use sparse representation to correlate the sparse edges and blur strength. Frequency-based methods are another noticeable trend in hand-crafted features, since the high-frequency components of the in-focus region and out-of-focus region are different. For example, Golestaneh~\textit{et al.}~\cite{Golestaneh:2017tv} use multi-scale high-frequency fusion and sort transform to determine the magnitudes of gradients. Although the methods based on hand-crafted features have been demonstrated to be effective in some cases, these methods are not robust enough in a broader variety of complex scenes.

\noindent\textbf{Learning-based methods}
Deep neural networks, especially CNNs, are widely used in many computer vision and image processing tasks.  Park~\textit{et al.}~\cite{Park:2017uq} propose the first CNN based method to DBD by combining the hand-crafted features and pre-trained deep features together. In this method, the image is scanned in a patch-by-path manner to find the defocus blur. Inspired by the object detection and segmentation methods, Zhao~\textit{et al.}~\cite{zhao2018defocus,zhao2019defocus} firstly use the full convolutional network-based method by considering DBD to be sensitive to scale. Following this idea, Tang~\textit{et al.}~\cite{tang2019defusionnet} design a novel network structure for feature fusion and Zhao~\textit{et al.}~\cite{Zhao_2019_CVPR} ensemble multiple networks to enhance diversity. In contrast to previous studies, Lee~\textit{et al.}~\cite{LeeLCL19} address the lack of datasets by learning from synthesized rendered dataset with domain adaptation. However, previous learning-based methods only focus on learning with stronger networks\cite{Zhao_2019_CVPR,zhao2019defocus,zhao2018defocus} or dataset~\cite{LeeLCL19}.

\noindent\textbf{Depth estimation and depth-assisted methods}
Estimating the depth from a single image is ill-defined since inferring 3D information requires multi-views. However, monocular depth estimation in restricted scenes is possible, for example, with indoor scenes~\cite{eigen2014depth} or the road in a driving context~\cite{godard2017unsupervised,godard2019digging}. In contrast, predicting the depth in the wild is still a challenge. Chen~\textit{et al.}~\cite{chen2016single} propose an end-to-end network based on point relationships. However, this network only predicts relations between the objects other than absolute depth. Li~\textit{et al.}~\cite{li2019learning} generate the multi-view disparity of humans from video of people who are frozen in place, and this task only works when the person are in the scene. Depth also plays an important role in other tasks. Most methods consider the depth to be known by the sensor. Such as RGB-D object detection~\cite{qi2018frustum} and RGB-D salient object detection~\cite{peng2014rgbd}. Some methods exploit the knowledge of depth in related tasks, such as  depth-assisted view synthesis~\cite{cun2018depth} and depth attentional features for deraining~\cite{hu2019depth}. 

\section{Methods}
We define DBD as a supervised pixel-wise binary classification problem. Rather than considering the defocus region as positive, we learn the opposite DOF~(in-focus) region as previously~\cite{zhao2018defocus,zhao2019defocus,Zhao_2019_CVPR}. Given the input image $I$ and the corresponding ground truth DOF region $M$, we construct a deep convolutional network $\Phi(\cdot;\theta)$ by feeding the image $I$ to generate the DBD maps $\Phi_{df}(I)$ and depth maps $\Phi_{dp}(I)$. Then, we optimize the parameters $\theta$ of $\Phi$ to minimize the the defocus metrics $L_{df}$ and depth metrics $L_{dp}$: 
\begin{equation}
\mathop{\arg\min}_{\theta} L_{df}(M,\Phi_{df}(I;\theta))+L_{dp}(\Re(I),\Phi_{dp}(I;\theta))
\label{eq:0}
\end{equation}
where $\Re$ is a pre-trained depth estimation network\cite{chen2016single} for depth distillation. Below, we introduce the details of depth distillation, network structure and metrics.

\begin{figure}[ht]
\begin{minipage}[b]{0.4\columnwidth}
\centering
\subfigure[Depth Distillation]{
\centering\includegraphics[width=\columnwidth]{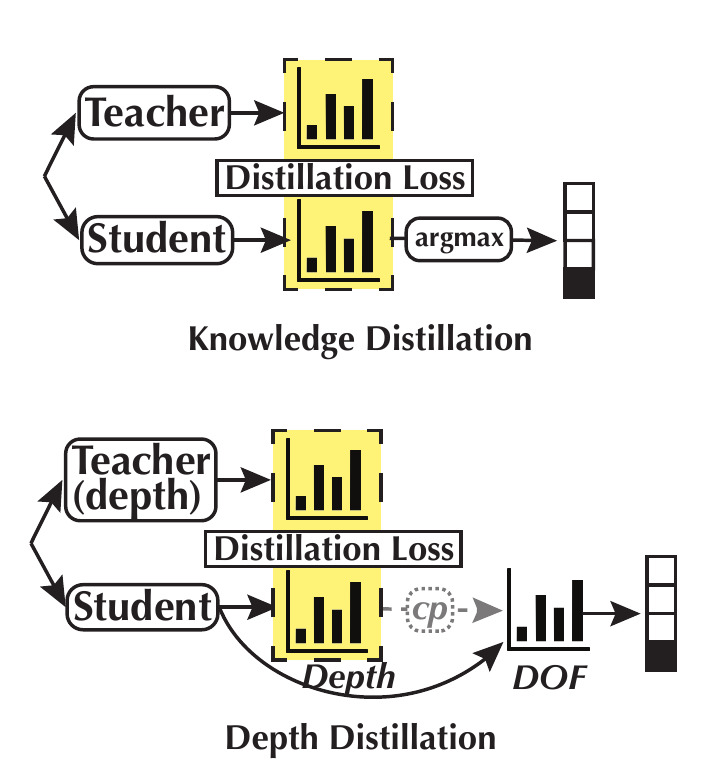}}
\end{minipage}
\begin{minipage}[b]{0.6\columnwidth}
\subfigure[Network Structure]{
\centering\includegraphics[width=0.9\columnwidth]{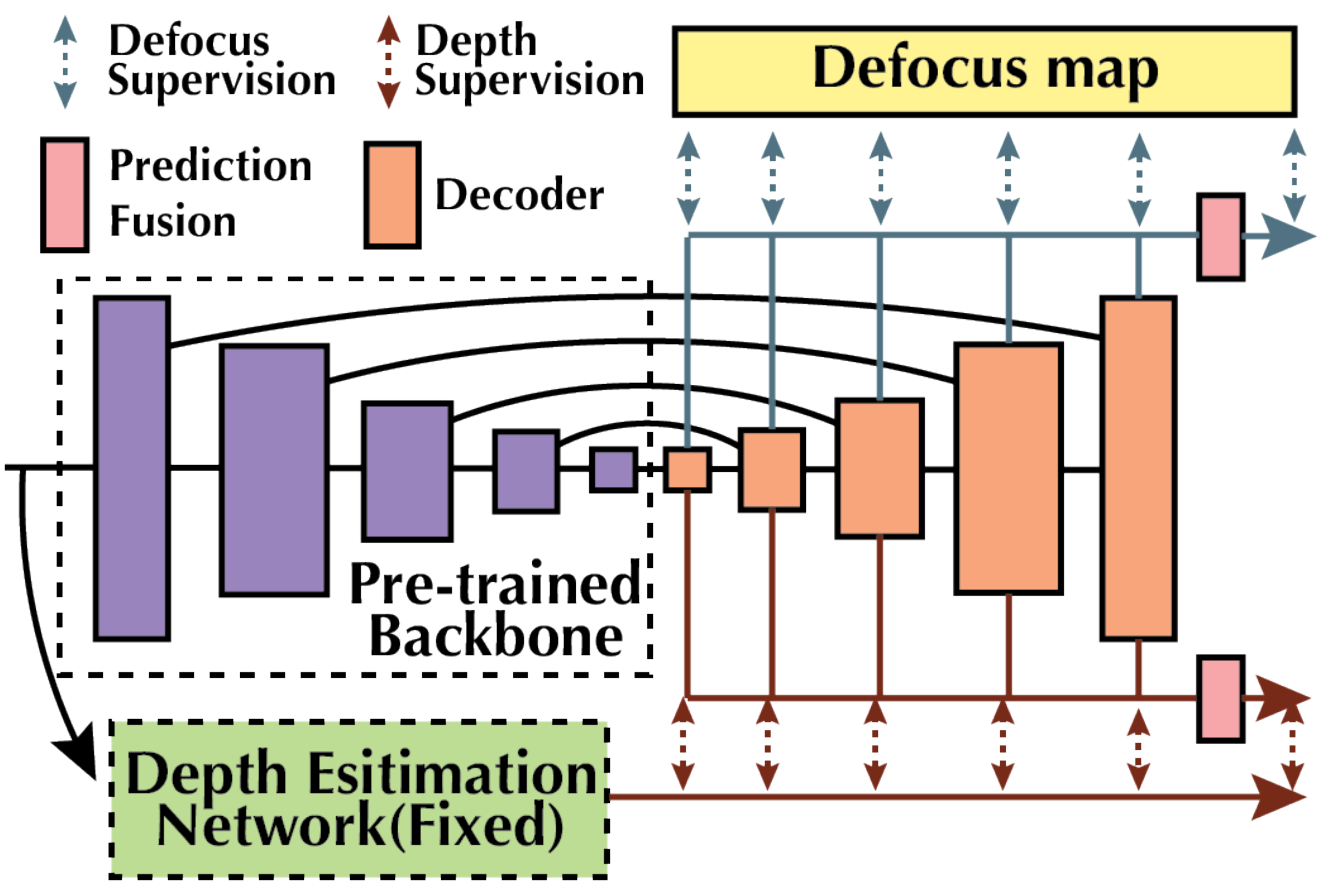}}
\end{minipage}
\caption{(a)Comparison with knowledge distillation and the proposed depth distillation. (b)Our network structure. Under FCN framework, we distill the depth information from a pre-trained depth estimation network\cite{chen2016single} and design novel decoders for DBD and depth estimation jointly.}
\label{fig:fcn}
\end{figure}

\subsection{Depth Distillation}
In general, knowledge distillation~\cite{hinton2015distilling,lopez2015unifying} aims to transfer the knowledge for network structure optimization. In detail, as shown in Fig.~\ref{fig:fcn}(a), they regularize the compact~(student) model using a larger~(teacher) network in the space of \textbf{continuous} soft label(the output of $\mathrm{Softmax}$), other than transferring the knowledge using predicted \textbf{discrete} hard targets.

Interestingly, we find that DBD(\textbf{discrete, classification task}) and depth estimation~(\textbf{continuous, regression task}) have a similar relationship with that between hard and soft labels in knowledge distillation. In photography, the sharp focus region~($DOF$) is mathematically defined as~\cite{dof}:
$DOF \approx \frac{2NCD^2}{f^2}$, where $N$ is the F-number of lenses, $C$ is the circle of confusion and $f$ is the focal length, respectively. The depth $D$ is the only one which is not the camera parameter~($cp$). 
Thus, as shown in Fig.~\ref{fig:fcn}(a), we propose \textit{depth distillation} to help defocus blur detection. In detail, we consider that the depth is the approximate soft label and distill the depth information from a pre-trained network as regularization of DBD. Instead of calculating the DOF from depth map directly and inferring the defocus map as knowledge distillation, our network can predict the defocus map and distill depth jointly because the camera parameters are unavailable. Although the structure of depth distillation and knowledge distillation are similar, the goal is totally different: We aim to involve the 3D information into DBD task other than distilling a compact model from teacher network. For implementation, we design a simple yet effective framework to achieve previous analysis. As shown in Fig.~\ref{fig:fcn}(b), we generate multiple outputs for depth estimation, which are supervised by a pre-trained network. Then, we fuse all the side outputs to obtain the final depth through a fusion~(1x1 Conv.) block. However, single image depth estimation is ill-posed since the dense depth is hard to be collected especially in unconstrained settings. Thus, we choose the relative depth network~(Chen \textit{et al.}~\cite{chen2016single}) as teacher network. Specifically, they aim to learn the relationships between scene objects other than accurate depth values. Thus, they label the spatial relationship between 800 pairs of points~(e.g., point A, B share the same depth, A is closer to camera than B and vice versa) pre-image as the supervision. Then, the neural network can predict the dense relative depth with the help of large-scale training samples. 

Leveraging depth information to DBD as depth distillation has many benefits. First, the depth distillation helps our network to understand the scenes better except for the binary classification~(similar to the relationships in knowledge distillation as discussed). Then, the blurriness region in the input also gives a dense hint to relative depth estimation. Finally, by depth distillation, we do not need the pre-trained depth network in testing, which also helps to build an efficient algorithm.
Distilling from relative depth network is also critical. Since the training dataset of DBD only contains 600 images, the pre-trained relative depth network(421K training images in the wild) involves more accurate 3D features from larger-scale datasets to our network and task. Besides, we find that the network of Chen~\textit{et al.} automatically locates the salient object and predicts its relative depth. Luckly, DBD has a similar goal because the photographers often use the defocus blur to stress the important views.
%
%

\subsection{Network Structure}
\begin{figure}[t]
  \centering
  \includegraphics[width=0.5\columnwidth]{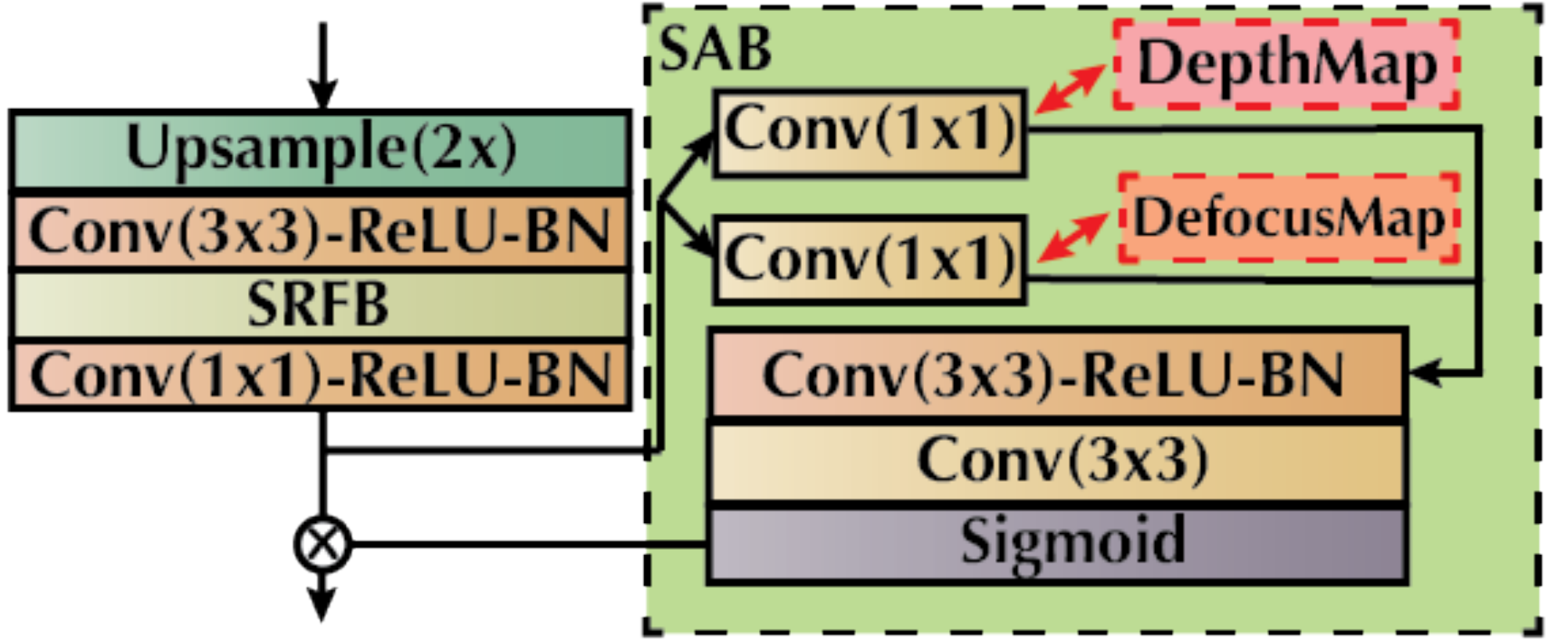}
  \caption{The detailed structure of the proposed decoder, where the red arrows mean defocus supervision and depth distillation, respectively.}
  \label{fig:umsampling}
\end{figure}

Our network structure is based on \textit{Fully Convolutional Networks}~(FCN~\cite{Long_2015_CVPR}). As shown in Fig.~\ref{fig:fcn}(b), we extract multi-scale features (5 layers in total) before each $\mathrm{MaxPooling}$ layer in a pre-trained ResNeXt101~\cite{Xie:2017un} on ImageNet. These multi-scale features contain both high-level semantic features and low-level details for further detection. In each decoder, as shown in Fig.~\ref{fig:umsampling}, we use the upsampling layer with convolution instead of deconvolution layer (or transpose convolution layer) to avoid checkerboard artifacts~\cite{LeeLCL19,odena2016deconvolution}. 
Then, by considering the importance of scale in DBD, we proceed using  several aspects of multi-scale feature modeling and preservation. 
On the one hand, we design auxiliary classifiers in each level of the decoder as in \cite{hou2017deeply,lin2017feature,LeeLCL19} to prevent over-fitting and to generate multi-scale results. Differently, in each level of the decoder, we design two auxiliary classifiers for the supervision from DBD and depth distillation, respectively. Each auxiliary classifier is defined as a 1x1 convolution layer for side prediction, and we reuse these side outputs as the Supervision-guided Attention Block~(SAB) for spatial attention ~(as shown in Fig.~\ref{fig:umsampling}). Then, the final defocus and depth map can be generated by merging all multi-scale intermediate output maps with a 1x1 convolution layer as the $\mathrm{Prediction Fusion}$ block in Fig.\ref{fig:fcn}(b). 
On the other hand, we model multi-scale reception fields in each level of the decoder and propose the Selective Reception Field Block~(SRFB) for efficiently selecting and merging the features in multi-contexts. Next, we provide the details of the proposed blocks.

\noindent\textbf{Supervision-guided Attention Block}
Inspired by recently proposed attention mechanisms~\cite{hu2018squeeze,woo2018cbam}, we increase the non-linearity of network with the attention block. In detail, we generate the attention map from the side outputs since it also has a stronger prior knowledge for further feature weighting. As shown in Fig.~\ref{fig:umsampling}, after the supervision of the auxiliary classifier, we feed the auxiliary outputs of DBD and depth to the network again. Then we generate the spatial attention by two convolution blocks and a $\mathrm{Sigmoid}$ function. Finally, we multiply the original features by the generated attention map. These attentions rescale the features spatially before the next decoder.

\noindent\textbf{Selective Reception Field Block} 
Since DBD needs to deal with scale carefully, previous works~\cite{zhao2018defocus,zhao2019defocus,tang2019defusionnet} merge multiple networks with multi-scale inputs, or recurrently and crossly fuse multi-scale features. However, these networks are still heavy and computationally inefficient. Rather than designing multi-stream networks or fusing by cross layers, we design an efficient multi-branch block for the extraction of multiple reception fields in each individual decoder.

\begin{figure*}[h]
  \centering
  \includegraphics[width=0.9\textwidth]{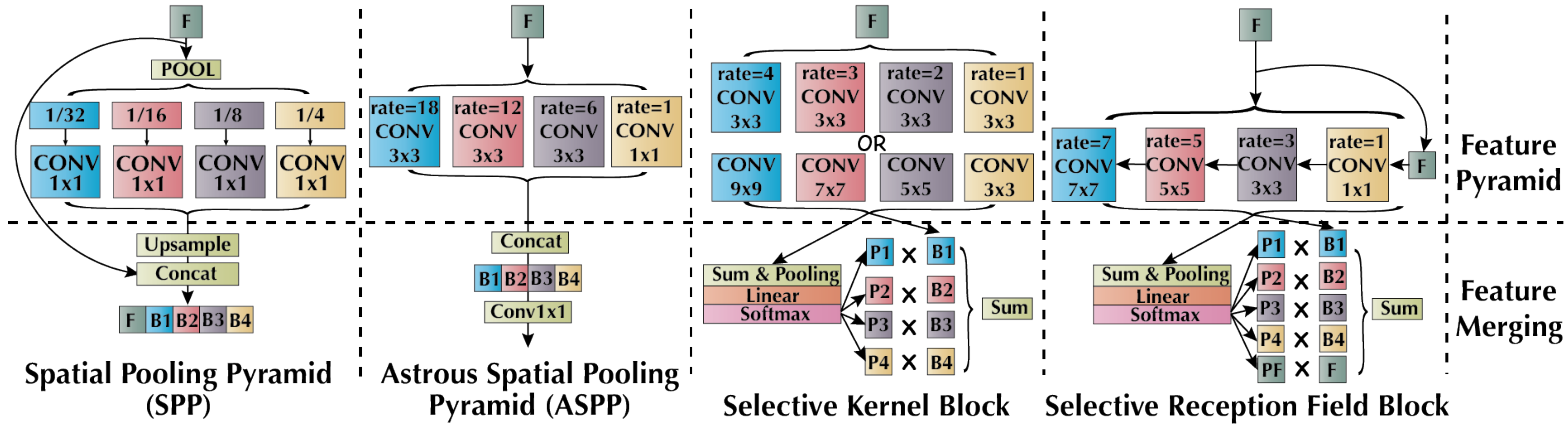}
  \caption{Different types of multi-kernel feature pyramids. The proposed Selective Reception Field Block enlarges the reception fields of Selective Kernel Block using larger reception field pyramids. $F$ repersents the original feature, $Bx$ and $Px$ are the $x$-th branch and the corresponding probability, respectively. $\mathrm{rate=k}$ means the dilation rate equals to $k$.}
  \label{fig:gsk}
\end{figure*}

This is a natural way to extract multi-scale features using different kernel sizes. For example, the widely used (atrous) spatial pooling pyramid~(SPP~\cite{zhao2017pyramid} or ASPP~\cite{chen2017deeplab}) has been successful in semantic segmentation and other related tasks~\cite{dehaze_zhang_2018,qu2019enhanced}. 
More recently, \textit{Selective Kernel Networks} (SK-Block~\cite{li2019selective}) have been proposed for weighting multiple kernels in image classification. As shown in Fig.~\ref{fig:gsk}, we find that the SK-Block has a similar purpose to (A)SPP. Thus, we give a general formulation of these blocks by modeling them as a two stage process: \textit{feature pyramid} and \textit{feature merging}. The (A)SPP extracts the multi-context features by pooling or dilated convolution and then merges with convolutional block. In contrast, the SK-Block models the multi-context features using different convolution kernels~(or convolution with different dilation rates). Then, it produces the probability of each branch using the global attention and uses it to weight each kernel. 


However, if we insert SK-Block to FCN directly, the reception fields are still local and enlarging it needs much more memory. Thus, inspired by the (A)SPP, we design the Selective Reception Field Block with the following improvements to SK-Block: First, we add the original feature into feature pyramid and merging. By involving the original feature, other branches will try to learn the residuals of input. On the other hand, we aim to create richer and larger receptive fields in the feature pyramid. In detail, we use a sequence of convolutional layers with the growth of dilation and convolutional kernel together as feature pyramid, which is inspired by \textit{Reception Field Block}~(RFB)\cite{liu2018receptive}. Note that RFB are proposed for object detection as an inception-like structure. In contrast, we build the feature pyramid, which is inspired by their intentions, and use these blocks as the decoders in the FCN framework. Thus, we have enlarged the reception field of the SK-Block substantially, which contains multi-scale features for weighting and selection. 
For example, when there are four branches in the block~(as shown in Fig.~\ref{fig:gsk}), the reception field of the Sk-Block is 11~($9\times9$ Conv. or $3\times3$ Conv. dilation=4) while ours is 43~($7\times7$ Conv. with dilation=7).

\subsection{Loss Function}
Our training loss is defined as a combination of overall auxiliary supervisions and the final prediction of defocus estimation and depth distillation. 

For defocus estimation, we use the weighed binary cross entropy~(BCE) loss as in~\cite{hu2018direction,Zheng_2019_CVPR} for all the auxiliary outputs $M'_{k}(k\in[1,..,5])$ and the final output $M_f$ compared with ground truth $M$:
$\ell_{defocus} = \ell_{bce}(M,M_f) + \sum_{k}\alpha^{k}\ell_{bce}(M,M'_{k})$, where the weighted BCE is defined as:
\begin{equation}
\begin{aligned}
  \ell_{bce}(M,M') = - (1-\frac{TP}{N_{p}})Mlog(M') -
   (1-\frac{TN}{N_n})(1-M)log(1-M')
\end{aligned}
\end{equation}
$TP$ and $TN$ are the numbers of true positives and true negatives in the  samples, $N_{p}$ and $N_{n}$ are the numbers of in-focus and out-of-focus pixels, respectively.

As for the depth distillation, giving the pre-trained depth estimation network as $\Phi_{rd}$, the input image $I$ and the predicted depth $\Phi_{d}(I)$ in our network, we define the depth distillation loss in level $k$ as:$ \ell_{depth}^{k} = ||\Phi_{d}^k(I) - \Phi_{rd}(I)||_2$.
Similar to defocus estimation, our network predicts multi-scale depth output and fuses the side outputs to generate the final results. Thus, the full loss of depth distillation can be written as:
\begin{equation}
  \ell_{depth} = \ell_{depth}^{f} + \sum_{k}\beta^{k}\ell_{depth}^k 
\end{equation}
where $\ell_{depth}^{f}$ represents the results after the final fusion layer and $\ell_{depth}^{k}$ represents the $k$ levels of auxiliary outputs.

Overall, the total loss of our network is:
$ L = \ell_{defocus} +  \gamma \ell_{depth}$. All the $\alpha, \beta$ are experimentally set to 1, and $\gamma$ equals to 0.1.

\section{Experiments}
\noindent\textbf{Implementation Details}
We implement our method in the PyTorch framework. The parameters of the encoder backbone are initialized from the pre-trained ResNeXt101~\cite{Xie:2017un} on ImageNet, while the other parameters are random noise. We utilize the Stochastic Gradient Descent~(SGD) algorithm to optimize the network with momentum of 0.9 and learning rate of 0.005. We resize all the images to 320x320 for training and evaluating the results in the same resolution as previous. Our network is trained on a computer equipped with an Intel 3.60 GHz CPU, 32G memory and a single GTX 1080 GPU. We set the batch size equals to 6, and the whole training process takes less than 2 hours. Regarding interference, our network can generate a 320x320 image in 0.028s~(\textbf{35.7 fps}), which is faster than previous DBD methods as shown in Table~\ref{tab:com}. Note that, for the training, we do not use any additional samples~\cite{zhao2019defocus} or synthesized samples~\cite{LeeLCL19}. Additionally, for fair comparison, all the results are raw outputs from the network without any post-processing~(such as dense conditional random fields~\cite{krahenbuhl2011efficient}). More comparisons and experiments can be found in the supplementary materials.

\begin{table}[t]
\begin{center}
\caption{Comparisons with state-of-the-art methods on $F^{\beta}$ and MAE score. We compare our method with 7 DBD methods~\cite{yi2016lbp,Park:2017uq,Golestaneh:2017tv,zhao2018defocus,zhao2019defocus,tang2019defusionnet,Zhao_2019_CVPR} and 4 methods on the related tasks(salient object detection~\cite{Qin_2019_CVPR,wu2019cascaded} and shadow detection~\cite{zhu18b,hu2018direction}). Our method achieves the best performance over 11 methods on two datasets. Meanwhile, our method is 2x faster than previous DBD methods.}
\label{tab:com}
\begin{tabularx}{\textwidth}{@{}clRRRRRRR|RRRR|R@{}}
\toprule
\multicolumn{1}{l}{Datasets} & Metrics &  \cite{yi2016lbp} & \cite{Park:2017uq} & \cite{Golestaneh:2017tv} & \cite{zhao2018defocus} & \cite{zhao2019defocus} & \cite{tang2019defusionnet} & \cite{Zhao_2019_CVPR} & \cite{hu2018direction} & \cite{zhu18b} & \cite{Qin_2019_CVPR} & \cite{wu2019cascaded} & Ours \\ \hline
\multirow{2}{*}{\begin{tabular}[c]{@{}c@{}}CUHK\\ 100\end{tabular}} & $F^{\beta}$ & .787 & .477 & .772 & .867 & .889 & .818 & .906 & .898 & .912 & .922 & .901 & \textbf{.927} \\
& MAE & .136 & .372 & .219 & .107 & .082 & .117 & .059 & .057 & .046 & .049 & .055 & \textbf{.042} \\ \hline
\multirow{2}{*}{\begin{tabular}[c]{@{}c@{}}DUT\\ 500\end{tabular}}  & $F^{\beta}$ & .719 & .468 & .687 & .761 & .827 & .823 & .817 & .844 & .877 & .827 & .866 & \textbf{.891} \\
& MAE & .193 & .410 & .248 & .194 & .138 & .118 & .135 & .109 & .080 & .120 & .092 & \textbf{.073} \\ \hline
- & FPS & .11  & .09  & .02  & .04  & .08  & 17.9 & 15.6 & 40.0 & 22.2 & \textbf{90.9} & 50.0 & 35.7 \\ \bottomrule
\end{tabularx}
\end{center}
\end{table}

\begin{figure*}[t]
\centering     
\subfigure[DUT500]{\centering\includegraphics[width=0.24\columnwidth]{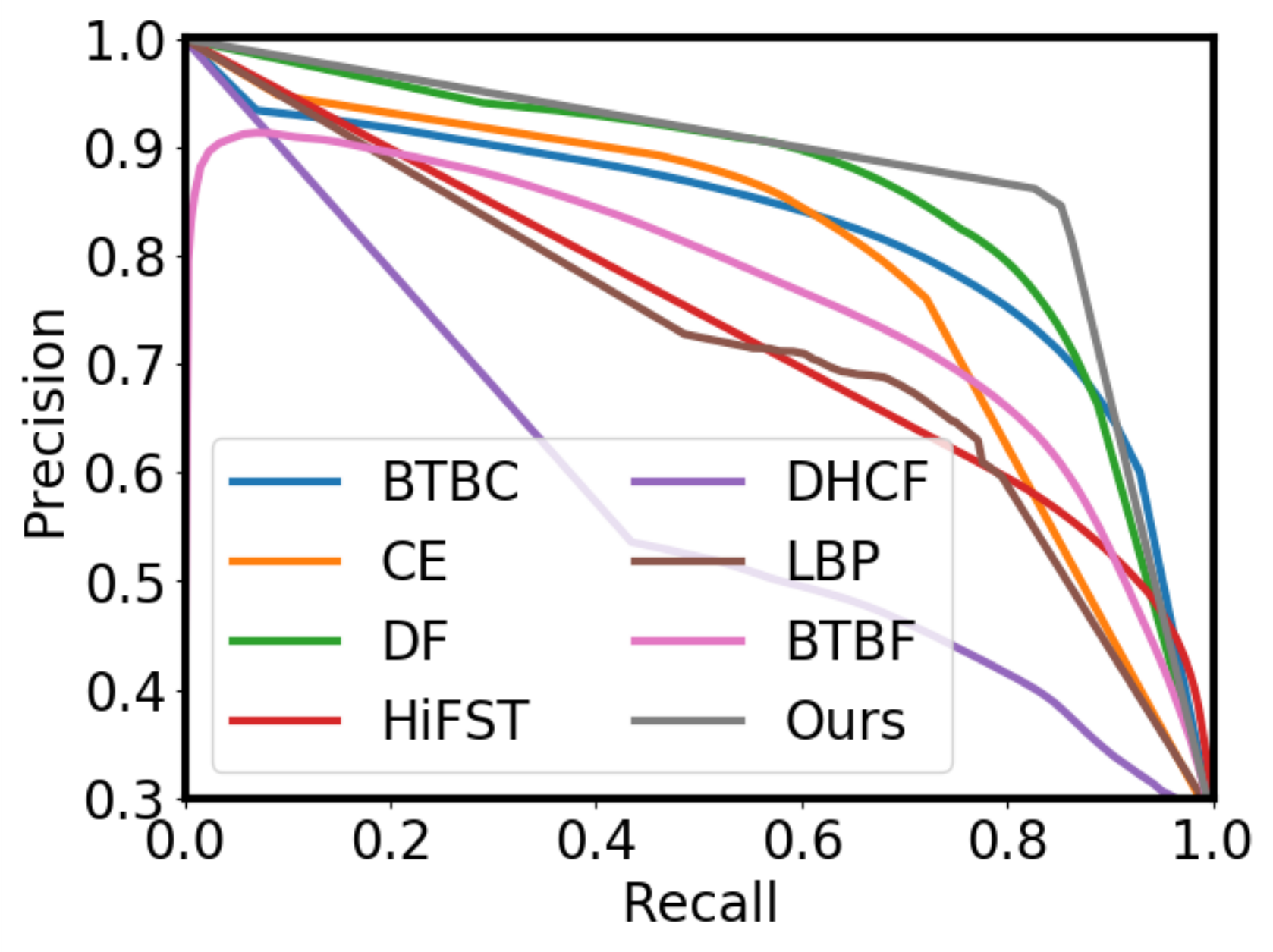}\label{fig:pra}}
\subfigure[CUHK100]{\centering\includegraphics[width=0.24\columnwidth]{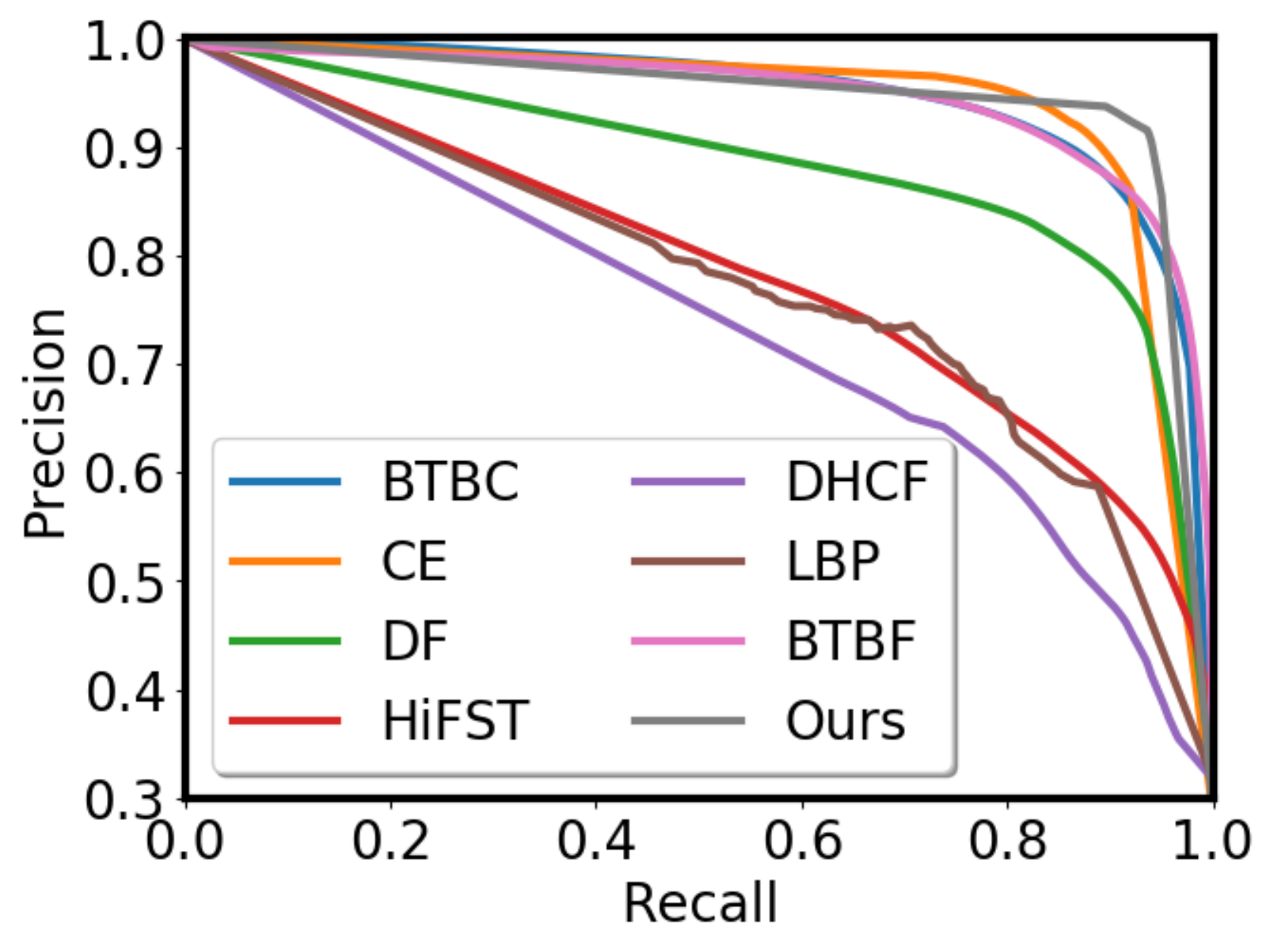}\label{fig:prb}}
\subfigure[DUT500]{\centering\includegraphics[width=0.24\columnwidth]{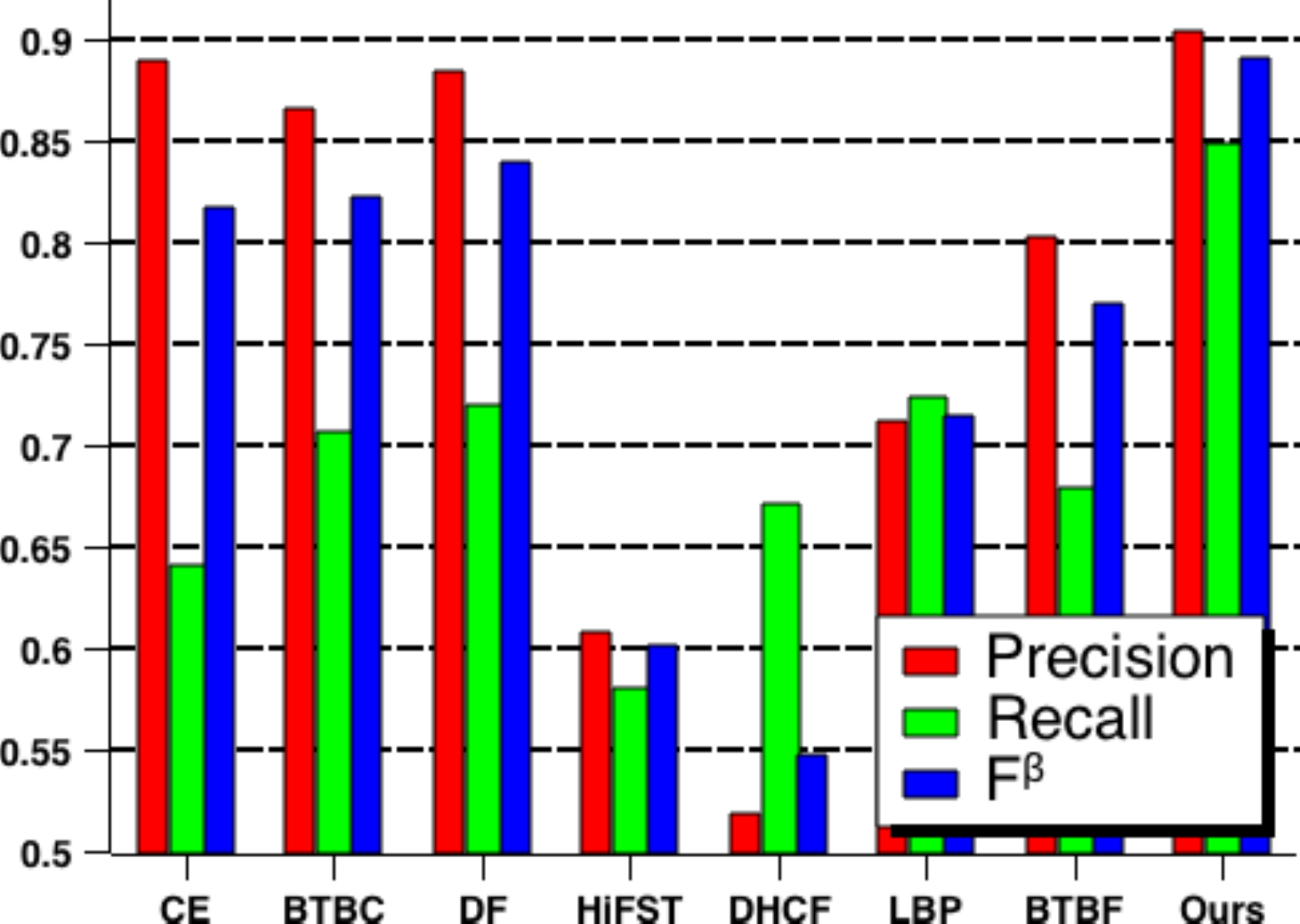}\label{fig:prd}}
\subfigure[CUHK100]{\centering\includegraphics[width=0.24\columnwidth]{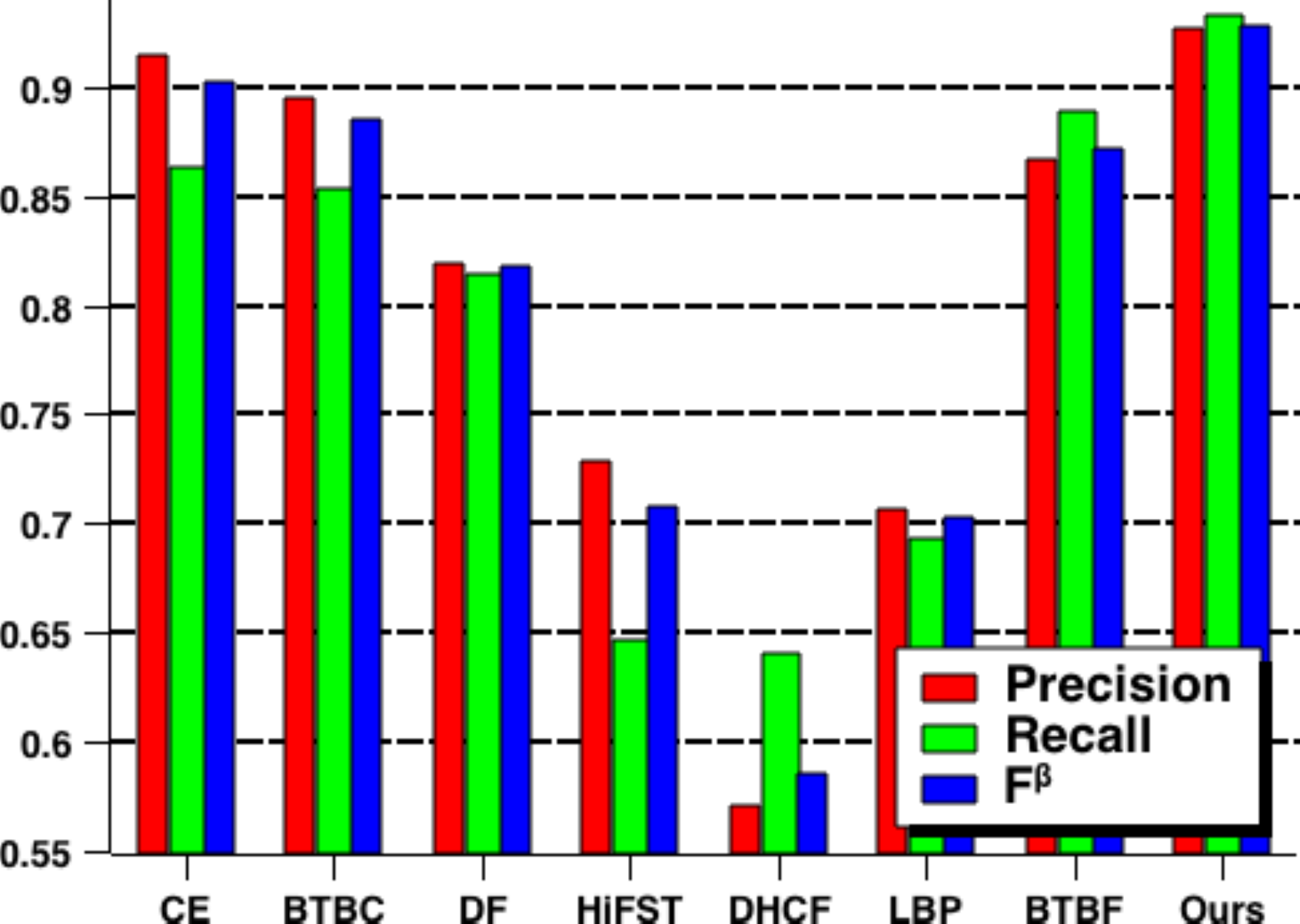}\label{fig:pre}}
\caption{ (a) and (b) are \textit{Precision-Recall Curves}, and (c) and (d) are the comparison of Precision, Recall and $F^\beta$ on two datasets. The proposed method achieves the best performance on various metrics.}
\label{fig:pr}
\end{figure*}

\begin{figure}[t]
  \includegraphics[width=\textwidth]{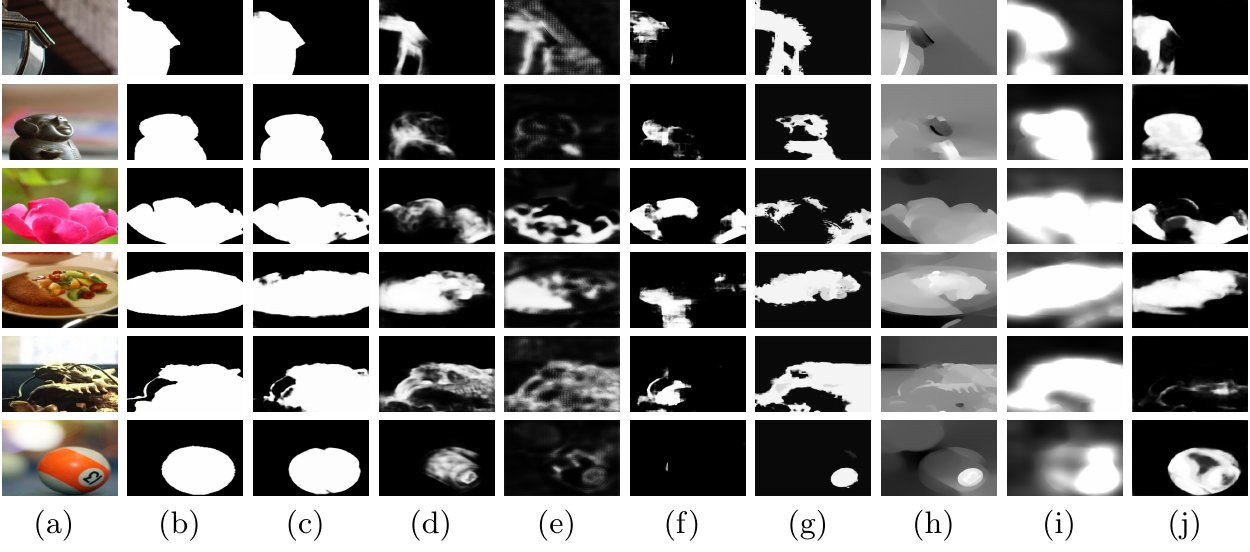}
\caption{Comparison with state-of-the-art DBD methods.  From the left to right is: (a)~Input, (b)~Target, (c)~Ours, (d)~BTB-C~\cite{zhao2019defocus}, (e)~BTB-F~\cite{zhao2018defocus}, (f)~CENet~\cite{Zhao_2019_CVPR}, (g)~LBP~\cite{yi2016lbp},(h)~DHCF~\cite{Park:2017uq},(i)~HiFST~\cite{Golestaneh:2017tv} and (j)~DFNet~\cite{tang2019defusionnet}. Our method generates more convincing DBD maps than others.}
\label{fig:novel}
\end{figure}

\noindent\textbf{Dataset}
We evaluate our algorithm on two publicly available datasets for DBD. The first is the CUHK dataset~\cite{Shi:2014vk}, which contains 704 images with partially defocus blur. Another dataset is the DUT dataset~\cite{zhao2018defocus}, which contains 500 difficult samples with obscure homogeneous, low-contrast focal regions and background clutter. We train our network on the same split of 604 images from the CUHK dataset as previous work~\cite{zhao2018defocus,zhao2019defocus,Zhao_2019_CVPR} and test on the remaining 100 images~(CUHK100) and the whole DUT dataset(DUT500). 

\noindent\textbf{Metrics} We evaluate DBD on three aspects as previous works. The first metric is the \textit{Precision-Recall (PR) Curve} for binary classification accuracy. All the results are normalized to [0, 255] and given a threshold in each integer interval. 
Second, we compute the mean precision, recall and F-measure scores~($F^{\beta}$) on the binarized results by an adaptive threshold. The threshold is determined by the 1.5 times of the mean pixel value. The F-measure is defined as:$ F^{\beta} = \frac{(1+\beta^2) \times Precision \times Recall}{\beta^2\times Precision  + Recall}$,
where $\beta^2 = 0.3$ and $Precision = \frac{TP}{TP+FP}$ and $Recall = \frac{TP}{FN+TP}$, respectively. A larger $F^{\beta}$ indicates a better result. Last, we report the mean absolute error~(MAE) for the average pixel differences between the ground truth $M$ and predicted $M'$. MAE is defined as: $MAE = \frac{1}{WH} \sum^{W}_{x=0}\sum^{H}_{y=0}|M(x,y)-M'(x,y)|$,
where $W,H$ are the width and height and $x, y$ are the spatial coordinates of the image, respectively.

\subsection{Comparisons with State-of-the-Art Methods}
We compare our algorithm with several state-of-the-art methods, including  deep learning-based methods for DBD, such as: deep and hand-crafted features based method (DHCF~\cite{Park:2017uq}), multi-stream bottom-top-bottom (BTB-F~\cite{zhao2018defocus}, BTB-C~\cite{zhao2019defocus}), network cross-ensemble(CENet~\cite{Zhao_2019_CVPR}) and the network with recurrently feature reuse and fusion~(DFNet~\cite{tang2019defusionnet}). In addition, we also conduct the experiments on state-of-the-art hand-crafted feature based methods, including local binary patterns (LBP~\cite{yi2016lbp}) and high-frequency multi-scale fusion and sort transform of gradient magnitudes (HiFST~\cite{Golestaneh:2017tv}). Note that, all the predicted maps of DBD come from the author's website or the public implementation with recommended hyper-parameters for fair comparison. For there are few learning-based DBD methods, we also compare our methods with 4 state-of-the-art learning-based methods on some relevant tasks:  
such as bidirectional feature pyramid network with recurrent attention (BDRAR~\cite{zhu18b}) and direction-aware attention (DSC~\cite{hu2018direction}) for shadow detection, boundary-aware loss~(BAS~\cite{Qin_2019_CVPR}) and cascaded partial decoder~(CPD~\cite{wu2019cascaded}) for salient object detection. All the networks of relevant tasks are trained on our framework with the same input resolution and batch size.

\begin{figure}[t]
  	\includegraphics[width=\textwidth]{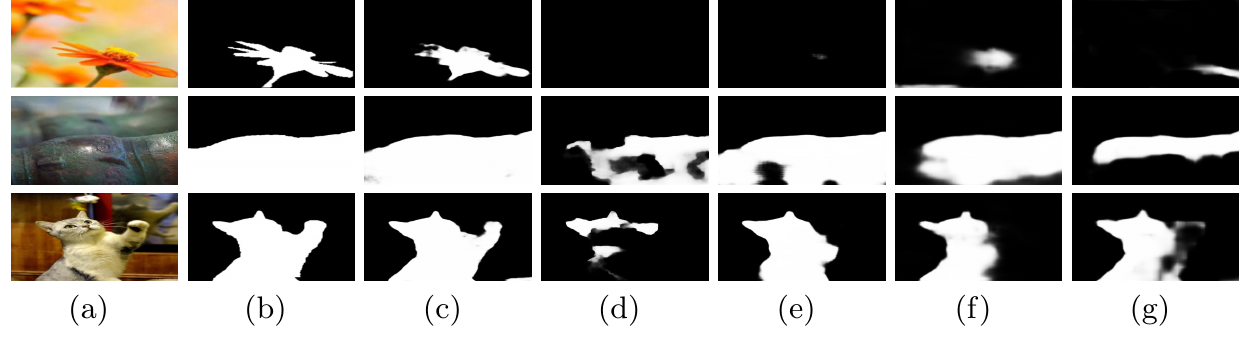}
	\caption{The produced DBD map of our method outperforms others state-of-the-art network structures on related tasks. From the left to right is: (a)Input, (b)Target, (c)Ours, (d)BASNet\cite{Qin_2019_CVPR}, (e)BDRAR\cite{zhu18b}, (f)CPD\cite{wu2019cascaded}, (g)DSC\cite{hu2018direction}.}
\label{fig:related}
\end{figure}

We illustrate the numerical comparison of our method and state-of-the-art methods on two public datasets in Table~\ref{tab:com} and Fig.~\ref{fig:pr}. It is clear that our method outperforms others with a larger margin on all numerical metrics. The results show that our network with depth and multi-scale features understands the complex scenes well. We also give some visual samples to compare with state-of-the-art DBD methods in Fig.~\ref{fig:novel}. Our methods also show the superior visual quality. Apart from the great object awareness in examples, our network also predicts the homogeneous regions well~(such as the plane in the fourth example) because of depth distillation. For comparison with related tasks, Table~\ref{tab:com} also gives a clear result. Our network has better performance in DBD than the boundary awareness network BASNet~\cite{Qin_2019_CVPR} or direction awareness DSC~\cite{hu2018direction} because depth is more important in our task. For example, boundary loss in BASNet~\cite{Qin_2019_CVPR} is benefit on CUHK100~(As Table~\ref{tab:com}) but worse in DUT500 because the homogeneous regions in DUT500 are not related to edge. As shown in Fig.~\ref{fig:related}, our method can achieve much better results than the other methods.

\subsection{Ablation Studies of Network Structure}

\begin{table}[t]
\begin{center}
\caption{Ablation study. The first two experiments use VGG19 as feature extractor while the last five experiments use ResNeXt101 as feature extractor. OursFull means the FCN+D+SRFB+SA.}
\label{tab:abl}
\begin{tabularx}{\textwidth}{@{}llRR|RRRRR@{}}
\toprule
Datasets & Metrics & FCN VGG & OursFull VGG & FCN ResNeXt & +D    & +D +SRFB & +D+RFB +SA & OursFull ResNeXt \\ \hline
\multicolumn{1}{c}{\multirow{2}{*}{\begin{tabular}[c]{@{}c@{}}CUHK\\ 100\end{tabular}}} & $F^\beta$ & 0.911 & 0.919 & 0.917 & 0.922 & 0.926 & \textbf{0.931} & 0.927 \\ 
\multicolumn{1}{c}{} & MAE  & 0.053  & 0.048 & 0.046  & 0.046 & 0.045   & \textbf{0.040}     & 0.042    \\ \hline
\multicolumn{1}{c}{\multirow{2}{*}{\begin{tabular}[c]{@{}c@{}}DUT\\ 500\end{tabular}}} & $F^\beta$ & 0.800 & 0.844 & 0.879 & 0.883 & 0.888 & 0.887 & \textbf{0.891} \\
 & MAE & 0.148 & 0.113 & 0.080 & 0.077 & 0.076 & 0.075 & \textbf{0.073}    \\ \bottomrule
\end{tabularx}
\end{center}
\end{table}

\begin{figure}[t]
\centering
\includegraphics[width=\textwidth,height=35mm]{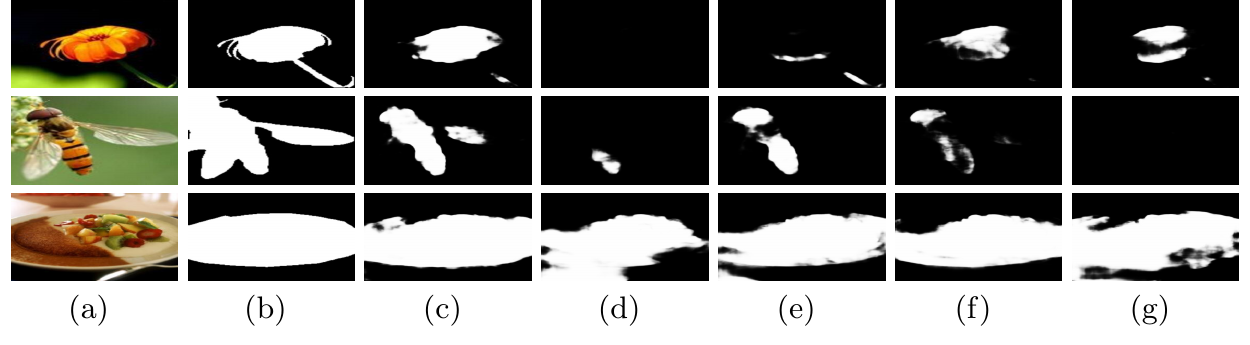}
\caption{Ablation study of network structure. From left to right is: (a) input, (b)~Target (c)~defocus+D+SRFB+SA (Our Full method) (d)~defocus only, (e)~defocus+D, (f)~defocus+D+SRFB, (g)~defocus+D+RFB+SA. }
\label{fig:abl}
\end{figure}

\textbf{Backbone choice} We choose different backbones for our network structure, especially the widely used VGG19 in previous work and ResNeXt101. For the ablation study, we use the FCN~\cite{Long_2015_CVPR} with auxiliary outputs and ResNeXt101 as feature extractor and compare with our main contributions  in Table~\ref{tab:abl}. Since the CUHK100 dataset is small and simple, the metric differences on this dataset is not too large. While on the DUT500 dataset, ResNeXt101 can extract richer features and gain much better results. By comparing with the other state-of-the-art methods on DBD~(7 DBD methods~\cite{yi2016lbp,Park:2017uq,Golestaneh:2017tv,zhao2018defocus,zhao2019defocus,tang2019defusionnet,Zhao_2019_CVPR} in Table~\ref{tab:com}, and 4 related tasks in Table~\ref{tab:com} ), our network can also improve the performance significantly on the similar pre-trained VGG19 backbone.

\noindent\textbf{Depth Distillation~(D)}
We test the effectiveness of depth distillation for DBD in Fig.~\ref{fig:abl}(d)(e) and Table~\ref{tab:abl}. It is clear that with the help of depth, our network can understand scene well and gain much better results because the depth information gives a strong prior for defocus map detection. Using Depth distillation, our network can also predict the relative depth from a single image. Although it is not our main target and our network can only predict the depth for partial defocus images, we still compare the distilled depth with our teacher network~(Chen~\textit{et al.}~~\cite{chen2016single}) in the supplementary materials.

\noindent\textbf{Depth Distillation Hyper-Parameter $\gamma$} We evaluate the influence of depth distillation hyper-parameters $\gamma$ on DBD. Thus, we train our full method with different $\gamma$ values. As shown in the Table~\ref{tab:gamma}, when $\gamma$ is too large or too small, the performance become worse. Our network gain the best performance when $\gamma$ equals to 0.1.

\begin{table}[t]
\begin{center}
\caption{Hyper-parameters $\gamma$ for depth distillation. The best and second best results are marked in bold and underline, respectively.}
\label{tab:gamma}
\begin{tabularx}{\textwidth}{@{}lRRRRRR@{}}
\toprule
Datasets & Metrics & $\gamma = 0.01$ & $\gamma = 0.05$ & $\gamma = 0.1 $   & $\gamma = 1 $ & $\gamma = 5 $ \\ \hline
\multicolumn{1}{c}{\multirow{2}{*}{\begin{tabular}[c]{@{}c@{}}CUHK\\ 100\end{tabular}}} & $F^\beta$ & \underline{0.9253} & 0.9208 & \textbf{0.9267} & 0.9231 & 0.9222  \\ 
\multicolumn{1}{c}{} & MAE  & 0.0438 & 0.0442 & \textbf{0.0424}   & 0.0434 & \underline{0.0430}     \\ \hline
\multicolumn{1}{c}{\multirow{2}{*}{\begin{tabular}[c]{@{}c@{}}DUT\\ 500\end{tabular}}} & $F^\beta$ & 0.8844 & \textbf{0.8919} & \underline{0.8909} & 0.8902 & 0.8818  \\
 & MAE & 0.0737 & 0.0729 & \underline{0.0727} & \textbf{0.0696}  & 0.0786  \\ \bottomrule
\end{tabularx}
\end{center}
\end{table}

\begin{figure}[t]
  \includegraphics[width=\textwidth,height=35mm]{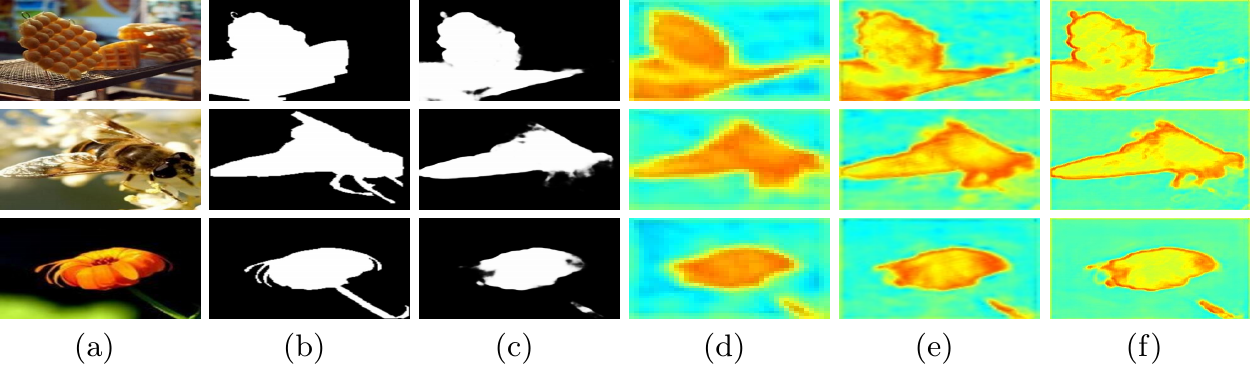}
  \caption{Visualized attention maps in SAB. From left to right: (a)~input, (b)~target, (c)~our final prediction, where (d)-(f) are the three attention maps in different levels of decoders. Here, we resize all the attentions to the same size for comparison. Interestingly, in high-level attention~(d), our SAB generates the attention map for the whole defocus region, while in the coarser level~(f), our attention map focuses on learning the edges and details.}
\label{fig:att}
\end{figure}

\noindent\textbf{Selective Reception Field Block~(SRFB)} We evaluate the performance of the proposed SRFB by inserting the SRFB in each level of the decoder. As shown in Fig.~\ref{fig:abl} and Table~\ref{tab:abl}, the SRFB models multi-scale features from the input and generate more accurate results. In addition to the necessity of our SRFB shown in Table~\ref{tab:abl} and Fig.~\ref{fig:abl}, we also conduct the experiments to compare our SRFB with the model without selective attention~(similar to FRB~\cite{liu2018receptive}). As shown in Fig.~\ref{fig:abl} and Table~\ref{tab:abl}, although the FRB perform better in the CUHK100 dataset, our SRFB show a much better results in DUT500. We argue that CUHK100 is smaller and easier. Thus, the proposed SRFB is more suitable for DBD.

\noindent\textbf{Supervision-guided Attention Block~(SAB)} In each level of auxiliary outputs, we design SAB to reuse the predicted defocus and depth map as spatial attention for further prediction. These attentions emphasize the useful features for further refinement. As shown in Fig.~\ref{fig:abl}(c)(f) and Table~\ref{tab:abl}, the proposed SAB also benefits blur detection. We also plot different levels of attention maps in the proposed SAB in Fig.~\ref{fig:att}. It is shown that using side outputs to generate the attention map emphasizes different features in each of their scales. Higher-level attentions stress the global features while the lower ones focus on local details.

\subsection{Failure Cases}

\begin{figure}[t]
\centering
  \includegraphics[width=\textwidth]{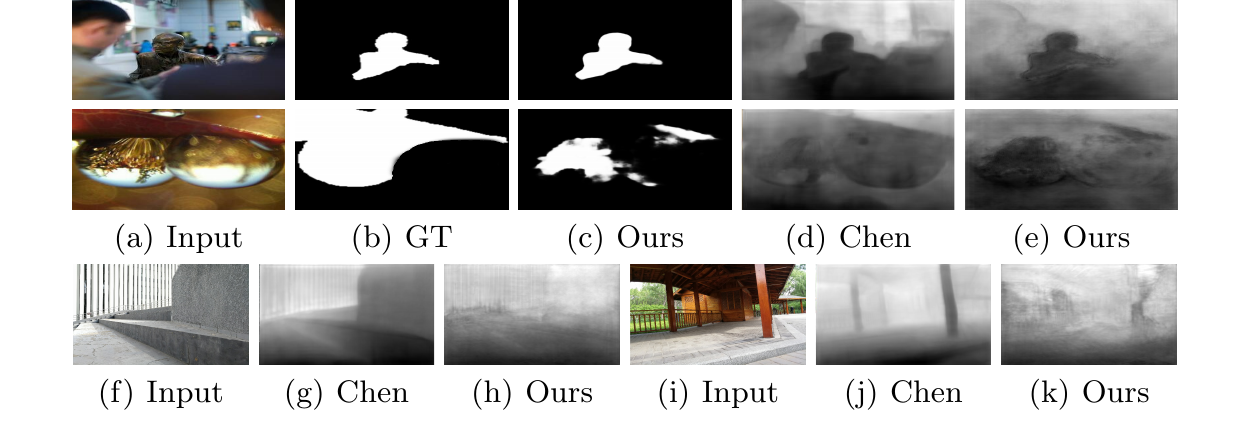}
  \caption{Failure cases. The top two rows show the failure examples when the scenes are complicated or when the depth is hard to predict. The thirds line shows the failed cases of predicted depth for all-in-focus image.}
  \label{fig:fail}
\end{figure}

Although our network shows much better results than previous methods, there are still some failure cases. As shown in the first row of Fig.~\ref{fig:fail}, when the far and near out-of-focus regions appears in a single image, the proposed network successfully predicts the defocus map but the relative depth relationship of the front person is incorrect. We also plot a more complicated example in the second row, the depth in this scene is hard to estimate because of the reflection of the water drop. Therefore, the proposed network cannot obtain global information and only predicts the scenes in the water drop. However, we argue that these problems can be mitigated by stronger networks and larger datasets.

Another limitation is our depth estimation. Our network can only predict the relative depth for partially defocused images, not depth estimation in the wild as in Chen \textit{et al.}~\cite{chen2016single}. We randomly choose two all-in-focus images and plot the results in the third line of Fig.~\ref{fig:fail}. When the image is all-in-focus, the defocus maps will not provide an effective prior for depth estimation. Thus, the apply range of our depth estimation is limited. However, our main target is DBD other than depth estimation.

\section{Acknowledgments}
The authors would like to thanks Nan Chen for her helpful discussion. This work was partly supported by the University of Macau under Grants: MYRG2018-00035-FST and MYRG2019-00086-FST, and the Science and Technology Development Fund, Macau SAR (File no. 041/2017/A1, 0019/2019/A). 

\section{Conclusions}
In this paper, we firstly discuss the role of depth in defocus blur detection and propose depth distillation for this task. In detail,
we distill the relative depth as regularization for learning-based defocus blur detection in a FCN network. Moreover, in order to build a stronger network, we design a selective reception field block because DBD is sensitive to multi-scale features, and we design a supervision-guided attention block, which serves the side outputs as spatial attention. The experimental results show the superiority of our method compared with 11 state-of-the-art methods in terms of efficiency and accuracy. 

\clearpage
%
%
\bibliographystyle{splncs04}
\bibliography{egbib}
\end{document}